\documentclass[a4paper]{article}
\usepackage{amsmath}
\usepackage{graphicx}
\usepackage{soul}
\usepackage{cite}
\usepackage{float}
\usepackage{amsfonts}
\usepackage{amsthm}
\usepackage{enumitem}

\title{ What is Learning ? \\
\large A primary discussion about Information and  Representation}
\author{Wu Hao\\
        email: wuhao29@gmail.com ; hw1034@york.ac.uk\\
        Department of Computer Science, University of York}

\begin{document}
    \maketitle
    \thispagestyle{empty}
    \pagenumbering{gobble}
    \pagenumbering{arabic}
    
    \newpage
    \tableofcontents    
    \thispagestyle{empty}

    \newpage

\begin{abstract}
Nowadays, represented by Deep Learning techniques, the field of machine learning is experiencing unprecedented prosperity and its influence is demonstrated in academia, industry and civil society. ``Intelligent" has become a label which could not be neglected for most applications; celebrities and scientists also warned that the development of full artificial intelligence may spell the end of the human race. It seems that the answer to building a computer system that could automatically improve with experience is right on the next corner. While for AI or machine learning researchers, it is a consensus that we are not anywhere near the core technique which could bring the Terminator, Number 5 or R2D2 into real life, and there is not even a formal definition about what is intelligence, or one of its basic properties: Learning. Therefore, even though researchers know these concerns are not necessary currently, there is no generalised explanation about why these concerns are not necessary, and what properties people should take into account that would make these concerns to be necessary. In this paper, starts from analysing the relation between information and its representation, a necessary condition for a model to be a learning model is proposed. This condition and related future works could be used to verify whether a system is able to learn or not, thus enriched our understanding of learning: one important property of Intelligence.
\end{abstract}

\thispagestyle{empty}
\newpage

\setcounter{page}{1}
\section{Introduction}
Although machine learning or artificial intelligence researchers know that we are not anywhere near the technique which could enable us to build a machine with real intelligence, almost all discussion about various properties of Intelligence stay on a philosophical level, such as the ability of learning, the ability of logical reasoning, self-consciousness, or even emotion. Formal mathematical definitions of these properties are very undefined, and we do not even have any binary criteria to verify whether a system really possesses any of these abilities, let alone formal methods which can be used to analyse the level of each of these abilities.

By generalising our intuitive understanding of learning ability, this paper gives a concise necessary condition for being a learning model. And this paper also presents further research directions which would enable us to analyse whether a model will be a learning model or non-learning model during the development process.

The discussion starts with intuitive introductions that prepare readers with conceptual understanding of the viewpoint of the following formal discussion.

\subsection{Intuitive introduction: Information}

We are surrounded and processing various information constantly. And certainty plays a critical role in representing information. Because one piece of information must be carried by at least one certain representation.
\begin{itemize}
\item ``I'm 188cm" : in this example a certain number together with a unit (cm) represents the information on height, the first person pronoun represents an identity related to the height information.
\item ``Turn left ? Right!" : in this example, the uncertainty of context brings ambiguity.
\end{itemize}

Even when we are measuring the uncertainty of a system (entropy), we will get a certain number or a range with certain boundary; or if we cannot get a certain boundary, we still have the certain representation of the system we are measuring; or even if we do not know what system we are measuring, we still have a certain definition of entropy; and if we have no certainty about anything, then there is no information at all. In other words, the amount of certainties also equivalent to the amount of information. Therefore, we can summary \textbf{Feature One} of information as follows:

\begin{center}
\begin{flushleft}
\textbf{One piece of information is equivalent to at least one certain representation, and vice versa.}
\end{flushleft}
\end{center}

\subsection{Intuitive introduction: Learning}

We as human cannot sense vast amount of physical realities naturally, such as the existence of almost all submicroscopic particles, the existence of majority region of electromagnetic wave, the movement of earth, the fact that the Earth is a sphere, actually this list could keep growing and includes almost all knowledge we have learned since the born of modern science. And we are all very familiar with the process of learning new knowledge based on known knowledge, and based on new learned knowledge to learn even newer knowledge. Firstly, this paper will give a generalised definition of the relative relation between the known knowledge and new learned knowledge. The introduction of this definition starts with an example:

\begin{center}
\begin{flushleft}
A man was charged with murder, judge or juror has no information about the reality of innocent or guilty. A reasonable sentence must be the consequence of implementing proper methods on all evidences.
\end{flushleft}
\end{center}

The evidence is \textbf{local information}, and the reality of innocent or guilty is \textbf{global information}. So the same as the known knowledge (\textbf{local information}) and new learned knowledge (\textbf{global information}) as introduced above. Actually, the notion of ``\textbf{no observation independent reality} " has been widely accepted for decades. The most important thing is, the global and local relation not only exists in learning high level abstract concept, but also exists in learning very fundamental concepts, such as object identification. Because we have no direct access to the world\cite[p.18]{luft2011subjectivity} other than though our sensors. The information of the existence of an object (\textbf{global information}) is the consequence of learning based on its different possible appearances (\textbf{local information}) received by our retina. And \textbf{Feature One} of information tells us there must be a certain representation of this object (\textbf{global information}).

Further discussion of learning (section 3) is after a formal discussion about information (section 2). These sections are all based on the explanation of mapping relations from a new viewpoint.

\section{Information}

\subsection{Information Generator and Representation}
\begin{figure}[H]
    \centering
    \includegraphics[width=\linewidth]{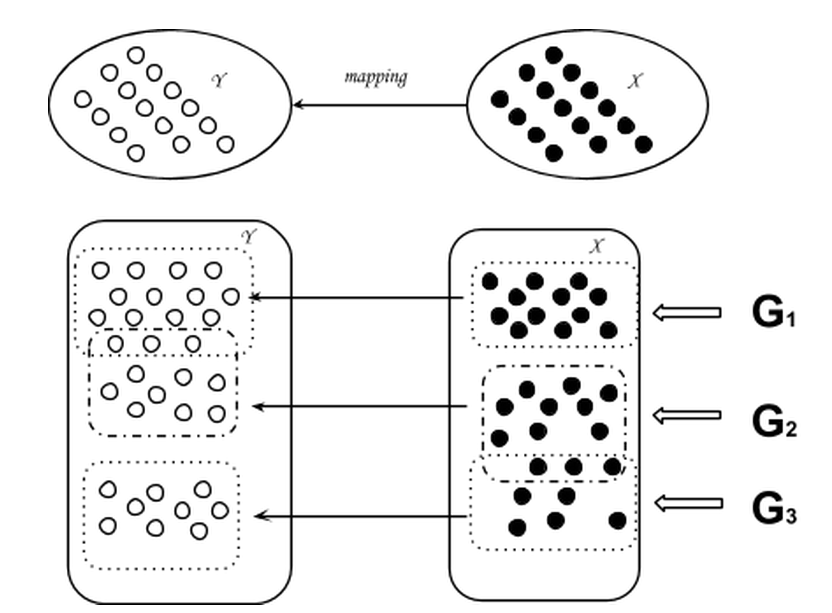}
    \caption{Mapping Relation and Generators}
    \label{p2}
\end{figure}

Suppose there is a mapping relation between a set $X$ and a set $Y$, so any subset of $X$ can be regarded as being generated by a generator, and the corresponding elements of subset of $Y$ are representations of this generator. The joint of any two subsets of $X$, or any two corresponding subsets of $Y$ are not necessarily to be null.

\subsection{Domain and Range of a Mapping Relation}
When a mapping relation $F$ is defined between domain $D$ and range $O$. Then, with respect to $F$, any subset of its domain could correspond to an unknown generator, and the corresponding elements in the range of that domain are the representation of that unknown generator.
\begin{figure}[H]
    \centering
    \includegraphics[width=\linewidth]{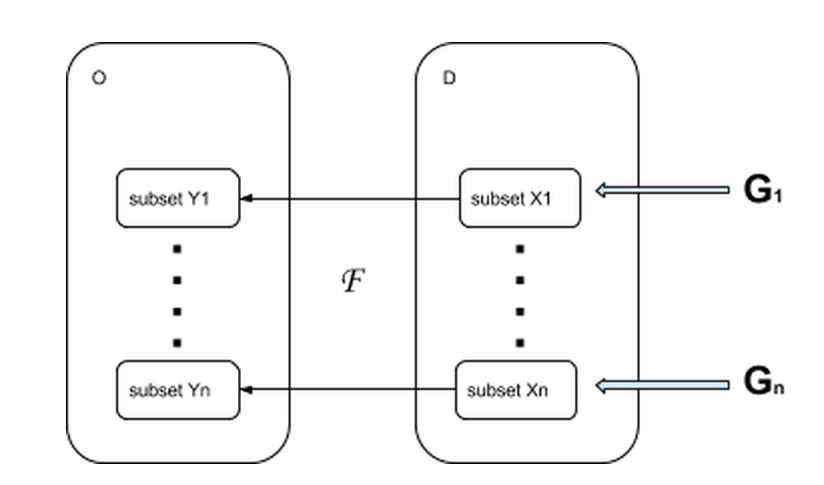}
    \caption{Mapping relation between domain and range}
    \label{p3}
\end{figure}
Then, with respect to $F$ (could be a function), we can define local information and global information as follows:
\begin{itemize}
\item Local information: Any input $x \in D $ is a local information with respect to $F$.
\item Global information: Any element of subset $Y \subset O$ and the corresponding subset $X = \{x| y \in Y, y = F(x)\}$ are global information.
\end{itemize}
So the global information $y \in Y, Y_n \subset O$ represents an unknown generator $G_n$. It is obvious that an unknown information generator $G_n$ could have a lot global representations: all elements of $Y_n$, and due to the \textbf{Feature One} of information, we can say that all elements of $Y_n$ should follows a certain constraint \textbf{$c$}. And if the corresponding elements of domain subset \textbf{$X_n$} do not follow the same constraint \textbf{$c$}, we can say that the global representation ($y \in Y_n$) of the information generator $G_n$ is the \textbf{invariant representation} with respect to its appearances ($\forall x \in X_n$).

\begin{center}
\begin{flushleft}
Example One:

\begin{itemize}
\item Domain: A CCD array, or our retina all provide a very large domain. If the value of each pixel is between 0-255 and the size of the image is 200*200 pixels, than the corresponding size of the domain is $256^{40000}$, which is a very large number.
\item Local information: Every image of a dog, every possible moment captured by our retina, these are all local information.
\item Global information: The existence of the dog, and all its appearances are global information. But we can learn the invariant representation form local information, a camera cannot.
\end{itemize}
\end{flushleft}
\end{center}

\subsection{Information (generator) Type}
It is obvious that different mapping relation pairs different subsets between domain and range.
\begin{figure}[H]
    \centering
    \includegraphics[width=\linewidth]{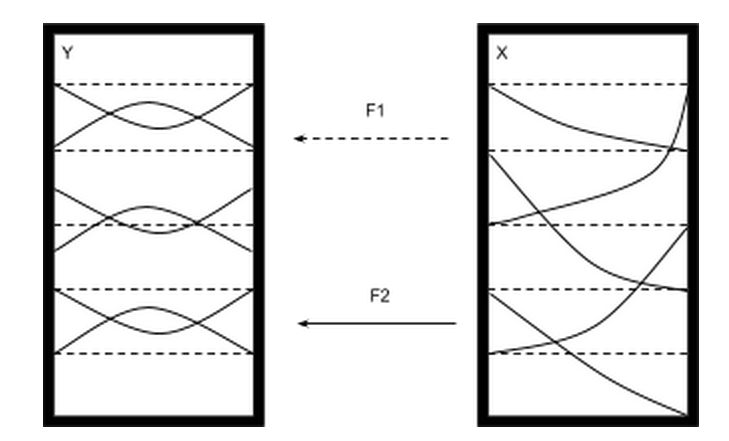}
    \caption{Different type of information}
    \label{p4}
\end{figure}
Therefore, we can say that a mapping relation along with its domain define the type of unknown generator $G_n$, or in other words defines the type of information.

\subsection{Common constraint:Linear separable hypersurface}
As introduced in section 2.2, global information must follow a certain constraint \textbf{$c$}, so the certainty of information can be guaranteed. There are a large number of possible constraints, and we denote the set of all possible constraint as \textbf{$C$}. Now we would like to know which constraint is preferable? and Why? The following discussion gives the answer of this two questions.

\begin{center}
\begin{flushleft}
Assumption One:

\textbf{Within certain type of information, there exists a mapping relationship that can define the differences between information generators.}
\end{flushleft}
\end{center}

Based on \textbf{Assumption One} and \textbf{Feature One} of information, we know that a mapping relation, denoted as \textbf{$F_C$}, map different subsets generated by unknown generators to global information that distinguishable by applying constraint \textbf{$c_R$}. By choosing the simplest constraint \textbf{$c_R$} that each $y \in O$ represent a distingusiable information generator, so the expression of  $F_C$ is:

\begin{center}
\begin{equation}
F_C: D \to O (D\subset R^N, O \subset R)
\end{equation}
\end{center}

\begin{center}
\begin{flushleft}
According to this expression, the differences of information generators can be identified as follow:

A set $X_1 \subset D$ is the global information of an unknown information generator $G_1$, then $y_1 = F_C(x): x \in X_1, y_1 \in R$ ; another set $X_2 \subset D$ is the global information of an unknown information generator $G_2$, then we have $y_2 = F_C(x): x \in X_2, y_2 \in R$.
\end{flushleft}
\end{center}
 If $y_1 \neq y_2$, then we can say that $G_1$ and $G_2$ are different information generators, and the type of information is defined by $F_C$.

\begin{figure}[H]
    \centering
    \includegraphics[width=\linewidth]{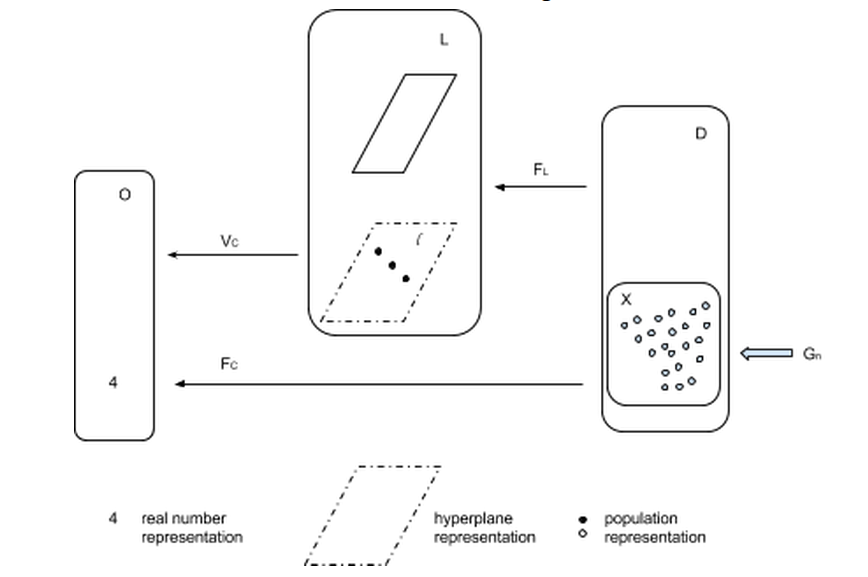}
    \caption{Representation and the Common Constraint}
    \label{p5}
\end{figure}

Furthermore, for real numbers in range $O$, they can be represent by parallel hyperplanes defined by a vector set $V_C$, so we can say that points on these hyperplanes are population representations (global information) \footnote{The term ``population representation in figure 4 means the invariant representation is high dimensional} of different unknown information generators, and this also explain why researchers prefer the property of linear separable so much, because it is not only easy understanding, but also is the common constraint that every element of constraint set \textbf{$C$} can convert to.

\section{Learning}

Previous section shows a formal way of explaining local information and its global representation. What being learned is nothing more than information, or in other words, is nothing more than global representation, so the analysis of learning ability should based on the analysis of the mapping relation as introduced in previous section. And again this section starts with an intuitive example.

\begin{center}
\begin{flushleft}

Example Two:
\begin{enumerate}[label =(\Alph*)]
\item select dob from table1 where name = ``wu hao"
\item $y = 2x$, when $x = 2$, we have $y = 4$
\item P(is human $|$ a picture of me) = 0.9999
\item People see picture 1, they know it is a pig. People see picture 2, they know it is a dragon. People see picture 3, even though they may have never seen this before and do not know the name, they still know it is different.

\begin{figure}[H]
    \centering
    \includegraphics[width=\linewidth]{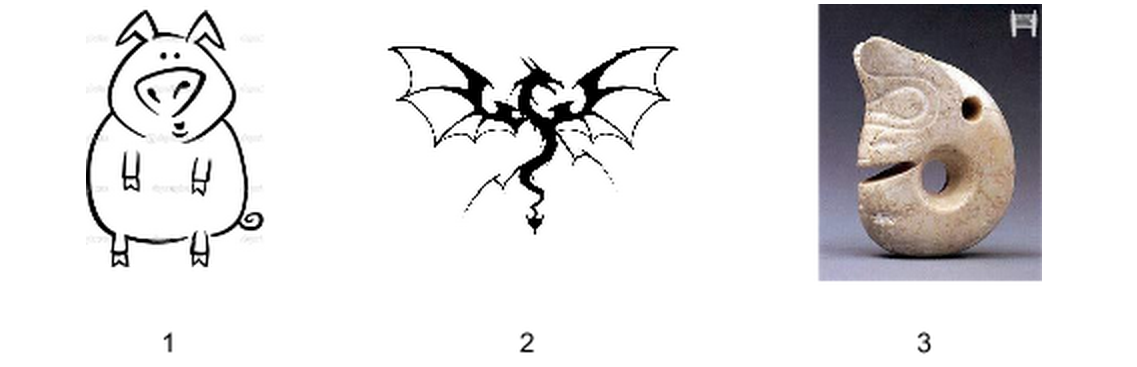}
    \caption{Three Pictures}
    \label{p1}
\end{figure}
\end{enumerate}
\end{flushleft}
\end{center}

These four models in example two all include one input, a mapping relation (function), and one output. A is a database, B is a linear function, C is a powerful probabilistic model, and D is us. By analysing example two from the mapping relation point of view we know that:

\begin{center}
\begin{enumerate}[label=(\Alph*)]
\item : The database will return $NULL$ for all inquiries that do not match what have been stored previously. 
\item : Outputs will be different when the inputs are different. And for almost all regression problems, we have assumptions about the mapping relation between two subsets of our observations, and after modelling our hypothesis of the mapping relation, for a new set of input we can almost always expect the output set contains new element which we have never seen before.
\item : When applying probabilistic model, our hypothesis is based on the belief that there are unknown statistical laws which represent different concepts \cite{vapnik2000nature} . Assume mapping relation $F_{model}$ is able to mimic the unknown statistical law of the appearance of a cup perfectly \footnote{This assumption helps to avoid the discussion of convergence problem.}, so the mapping relation we get is as follow:

\begin{figure}[H]
    \centering
    \includegraphics[width=80mm,scale = 0.5]{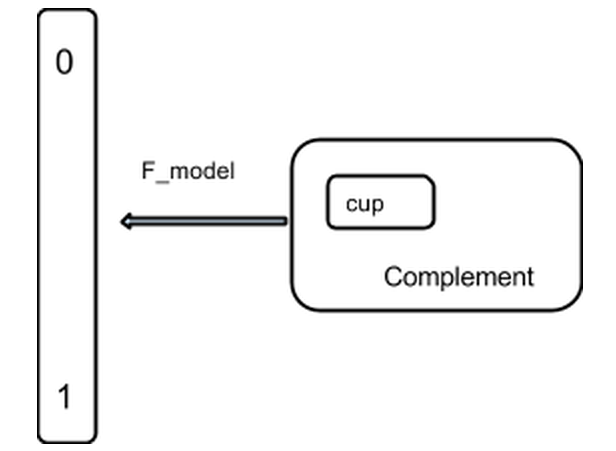}
    \caption{Probabilistic Model (example one)}
    \label{p8}
\end{figure}

At this stage, $F_{model}$ is able to give the chance of being a cup for any element of the domain. What if we want $F_{model}$ to have the ability to give the chance of being other object, such as a dog. Different from what we observed in regression problem, there is no real number naturally related to different objects. Therefore we chose another way of constructing our hypothesis of observation as follow:

\begin{figure}[H]
    \centering
    \includegraphics[width=80mm,scale = 0.5]{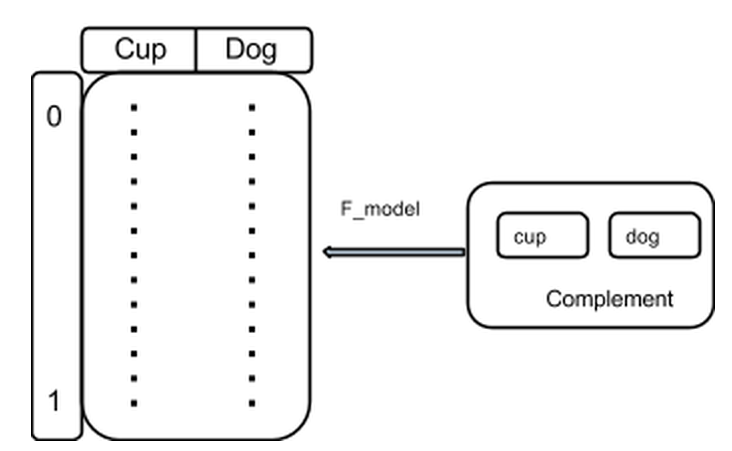}
    \caption{Probabilistic Model two}
    \label{p9}
\end{figure}

\begin{center}
$ F_{model}: R^N \to R \times I (I\ is\ indicator\ set) $
\end{center}

But the introduction of this ``Indicator Set" brings a problem. The size of this indicator set represent the information of the amount of objects $F_{model}$ can effectively identify. This is a global information (denoted as $\# I$) with respect to identities of each object, and it is given by us. For objects which are not included in our hypothesis of $F_{model}$, the outputs are almost all clustered around 0.

\item We, as human with well developed vision, are always able to identify object which have never been seen before, or in other words, the function of our vision system are able to define new global information.

\end{enumerate}

\end{center}

\subsection{Necessary condition of Learning}
Therefore, combining the discussion of information and its representation with above discussion, the intuitive understanding of the ability of learning is:

\begin{center}
\textbf{Learning means the ability to define new information generators}
\end{center}

From this point of view, B and D are learning model, A and C are non-learning model. And by generalising the description above, we have a formal necessary condition of a mapping relation $F_{model}$ to be a learning model as follow:

\begin{center}
\begin{flushleft}
Condition S:
\end{flushleft}

$ \forall X_S \subset D, Y_S = F_{model} (X_S), \exists X_N \subset D\setminus X_S : Y_N = F_{model}(X_N), (Y_N\bigtriangleup Y_S \neq Y_N)  \&  (Y_N\bigtriangleup Y_S \neq Y_S) $
\begin{flushleft}
For any subset $X_S$ of the domain D, the corrsponding subset of range is $Y_S$, exists a set $X_N$ which is a subset of the complement set of $X_S$, so that we have a subset $Y_N =F_{model}(X_N)$, the symmetric differences of $Y_N$ and $Y_S$ is neither $Y_S$ nor $Y_N$.
\end{flushleft}

\end{center}

Generally, if a mapping relation $F_{model}$ satisfies \textbf{Condition S}, then we can say $F_{model}$ is able to learn its domain and it is a \textbf{learning model}, otherwise it is not able to learn its domain and it is a \textbf{non-learning model}. But there are actually two strategies which enable a non-learning model $F_{model}$ to satisfy condition S, thus behaves like it is able to learn its domain.

\begin{center}
\begin{enumerate}[label=(\Alph*)]
\item Shrinking the scale of the domain.
\item Iterating over the domain
\end{enumerate}
\begin{flushleft}
\end{flushleft}
\end{center}

\begin{center}
\begin{flushleft}
Example Three:
\begin{figure}[H]
    \centering
    \includegraphics[width=\linewidth]{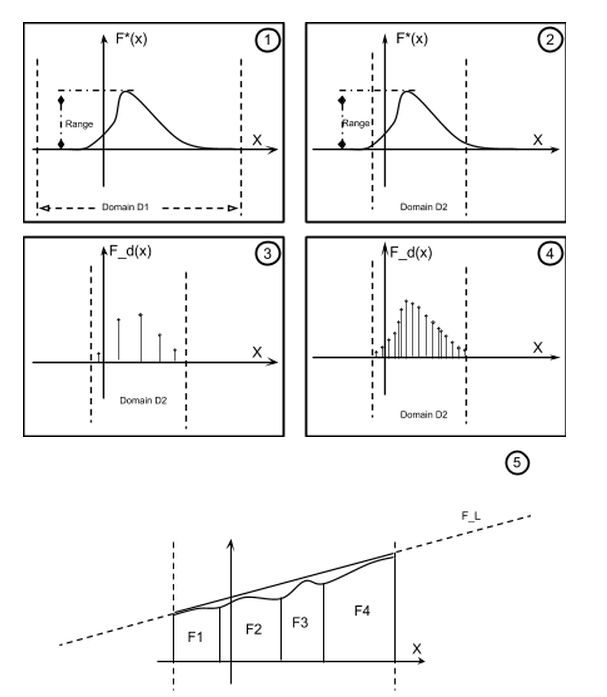}
    \caption{Strategies A and B}
    \label{p10}
\end{figure}

\begin{itemize}
\item As shown in figure 8, in picture 1, function $F^*$ is defined on domain $D$, and it is not able to learn its domain because it does not satisfy condition S. But by shrinking its domain as shown in picture 2, $F^*$ behaves like it is able to learn most of its domain. This is strategy A
\item As shown in figure \label{p10}, in picture 5, a set of function \{$ F_1, F_2,F_3,F_4$\} forms an approximation of function $F_L$ which satisfies condition S over its domain, but neither of these four functions satisfy condition S. This is strategy B.
\end{itemize}
Usually these two strategies are being applied together.
\end{flushleft}

\end{center}

For neutralising the effect brought by strategy A and B, condition S needs to be extended a little bit.

\begin{center}
\begin{flushleft}
\textbf{Condition $S^*$}:

Suppose domain D is defined on $N$ dimensional space, extend the scale of dimension $M$ of the domain to infinite and mapping relation $F_{model}$ still satisfies condition S, then we can say $F_{model}$ is a learning model on domain D, dimension M.
\end{flushleft}
\end{center}

If $F_{model}$ satisfies condition S* on every dimension of its domain $D$, then we can say $F_{model}$ is able to learn its domain $D$, and it is a complete learning model. On the other hand, as shown in figure 8 picture 5, the iteration process behaves similar with a database systems as shown in picture 3 and 4. And a database system is a typical memory system, so we can also say:
\begin{center}
\textbf{ 
\begin{itemize}
\item Not being able to satisfy condition $S^*$ is a sufficient condition for a model to be a Non-Learning model.
\item Non-Learning model and memory system are equivalent.
\end{itemize}
}
\end{center}

\subsection{AI Effect}
$F_{model}$ contains our hypothesis about the mapping relation between different subsets of our observation. The ideal situation is our hypothesis could be as close to the unknown mapping relation as possible. One method of measuring the validity of our hypothesis is as follow:

\begin{center}
\begin{flushleft}
Suppose unknown mapping relation is $U:D \to O$, and our guess if H. So for any $ o \subset O $, we have $T = U(o)$, and $G = H(o)$, then $J(T,G)$ \footnote{ $J(T,G)$ is the Jaccard similarity coefficient of these two sets, It can also being used to define the simplicity of a machine learning problem} is the validity of our hypothesis.
\end{flushleft}

\end{center}

For simple problems, such as regression problem, our hypothesis works fine , but its only learn simple mapping relation. But when facing complex optimisation problems, for getting a desired mapping relation $F_{model}$, strategies A and B are usually being applied together and if researchers do not exam the property of information being used carefully, it is possible that information comes from outside of this structure can be introduced. Usually the choice of introducing information from outside the structure will help to reduce the difficulties of constructing a desired $F_{model}$ or improve its performance. But this behaviour will eventually cause $F_{model}$ fail to satisfy condition $S^*$, as shown by the example of figure 7 (the probabilistic model).This somehow explains Rodney Brooks complain:

\begin{center}
\textbf{`` Every time we figure out a piece of it, it stops being magical; we say, 'Oh, that's just a computation "}
\end{center}

\section{Problems and Feature Works}
\begin{enumerate}
\item More necessary conditions or a sufficient condition

In this paper, condition $S^*$ is a necessary condition for a mapping relation to be a learning model, this does not rule out the possibility that some mapping relation $M_F$ could satisfy condition $S^*$ but still against our understanding of learning. Therefore, further discussion about condition on mapping relation is necessary.

\item Necessary condition on structural level

The discussion of previous section indicates that Condition $S^*$ of mapping relation is insufficient for analysing the structural detail of a given model, necessary condition on structural level is required to guide the construction process of getting a desired mapping relation and presumably would also give us a formal explanation about why introducing global information from outside the structure could make the problem easy to be solved or improve the performance but also causes losing the ability of learning at the same time.

\item Verifying state-of-the-art artificial intelligence systems.

Because of the overheated expectation of an realistic AI system and the following failure in 1970s, researchers usually avoid using the term ``AI" since then. Instead they express the similar idea by implying that their systems are able to learn similar features as human brain could do or be able to exhibit unexpected behaviours. 

These days, some systems have demonstrated impressive performances which not only make the discussion of AI a hot topic again but also raised the alert over the possibility of people may losing control of Artificial Intelligence. But no matter how powerful these systems could be, from machine learning point of view, they are still solutions of optimisation problems which based on different hypothesises of our observation. Therefore their ability of learning is independent from their performance and can be verified in the same way as the discussion of probabilistic model in section 3. The research results of problem one and two will provide more powerful tools for verifying the learning ability of a give model.

\item The common learning model

As discussed in section 2.4, linear separable is the common constraint that all constraints can convert to, therefore all learn models (include us) can convert to the common learning model which follows this common constraint. A prototype which demonstrates the common learning model theory is the zero to one step for harvesting information learned by future artificial intelligence system.

\item Begin with object recognition 

The ability of learning is one property of an intelligent system, the learning result is not necessary to be right all the time, and the concept of ``right" is also a global information which needs to be defined in future research. 

The process of learning highly abstract information is tightly related to other part of intelligence, after all, during the past 300000 years' development of homo sapiens, most of our learned highly abstract realities are ``Wrong" and comparing with realities we know today, it seems ``overfitting" is a common phenomenon of the highly abstract information learning process (World Elephant, then Geocentric Theory). Therefore, further study of learning will focus on the learning process of relatively independent and low level abstract information, such as the object recognition problem.

\item Dataset separation and merge problem

When explaining the object recognition problem using the theoretical framework proposed in previous sections, two seemingly counterintuitive deductions are dataset separation and merge problems.

\begin{itemize}
\item Dataset separation problem: When dataset generated by one object is separated , there could be two different set of invariant representations.
\item Dataset merge problem: When two dataset of different object are merged together, there could be a new set of invariant representations.
\end{itemize}

\begin{figure}[H]
    \centering
    \includegraphics[width=100mm,scale = 0.5]{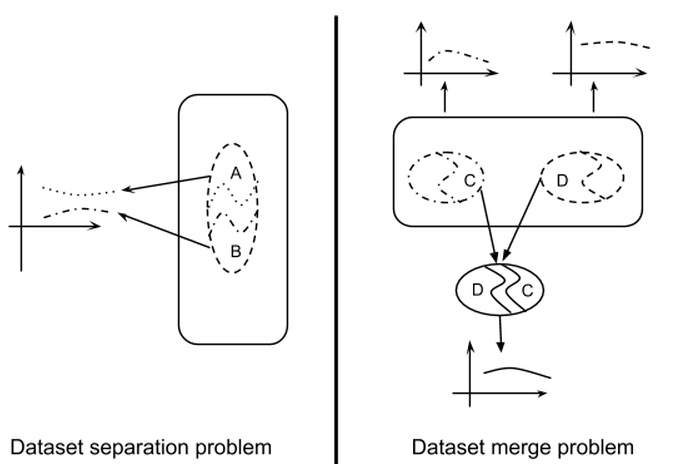}
    \caption{Dataset separate and merge phenomenon}
    \label{p12}
\end{figure}

The description above is the common intuitive understanding of these two problem, but there are two mistakes about this understanding:

\begin{itemize}
\item The information generators is defined by the mapping relation, so these two problems are equivalent.
\item Since there is no observation independent reality, this is not a problem, it is a phenomenon which can be used to verify the validity of future implementation based on this theoretical framework.
\end{itemize}

\begin{center}
\begin{flushleft}
\textbf{Note}

It seems particularly hard for people to understand the dataset merge phenomenon in object recognition. It indicates that if we could combine two objects together, such as a dog and a cat, and guarantee the smoothness of the transformation of their appearances, then we will see them as a same thing(things). This confused conclusion exists only because we have the information of the existence of these two distinguishable objects at first, and there is no real world experience which bring us into this mental experiment. 
But we can always almost recognise ``Optimus Prime", ``Bumblebee", ``Jazz" and other characters in transformers (the movie or the cartoon), no matter they appear as cars or human form robots. In this scenario, the recognition process of a transformer is no more difficult than recognise anything else and we take it for granted. Therefore, the confusion about cat and dog could be completely solved someday in the future which ``Marvel" or ``DC" decide to bring us a new superhero whose appearance transforms between cat and dog. And this phenomenon should be able to be verified experimentally if there are corresponding neuroscientific approaches.

\end{flushleft}
\end{center}

\item Other related topics

With the progress of future works, some other related topics will help to provide a better understanding about the essence of learning and its relation with intelligence. These topics include: detailed analysis of non-learning model, formal discussion of overfitting problem in the process of learning highly abstract information, strategies that could make a learning model behaves like a non-learning model.
\end{enumerate}

\section{Conclusion}
John Connor asked T-800:

\begin{center}
\begin{flushleft}
``Can you learn stuff that you haven't been programmed with? so you could be...you know, more human? And not such a dork all the time?"
\end{flushleft}
\end{center}

Even though that is just a scenario of a movie, still we all want to known the answer of this question. And for researchers, the most interesting part is how this question can be answered.

Machine Learning theories focus on solving different optimisation problems, so we could model hypothesises of our observation. While, the discussion of learning should focus on the behaviour of the model. In this paper, by analysing the relations among observable appearance(local information), reality (global information), their representation and the generalisation of our intuitive understanding of learning, we have a necessary condition $S^*$ which can be used to get the answer of John's question.

Starting from a viewpoint which has been missed out,
the analysis in this report shows that, Learning is not an action of absorbing existed information. In contrast, learning is an action of define information generators, or more specifically, is the ability of always being able to define new information generators. And preliminary verification indicates that our hypothesis of what we observed could be a learning model which only learn trivial information, and when facing complex problems \footnote{Complexity as defined in section 3.2.}, lack of learning ability is a trade-off for reducing the complexity or improving the performance through introducing global information from outside the structure.

It is possible in the future that we could harvest the information generated by a learning model which can learn like us but with greater learning ability. And focusing on the direction of getting a common learning model which can learn like us, this paper illustrates the logic dependencies of many topics that related to understanding the essence of learning.

\newpage

\bibliography{irl}{}
\bibliographystyle{plain}
\end{document}